%% file: main.tex
\documentclass[sigconf]{acmart}

\AtBeginDocument{%
  }

\setcopyright{acmlicensed}
\copyrightyear{2018}
\acmYear{2018}
\acmDOI{XXXXXXX.XXXXXXX}
\acmConference[Conference acronym 'XX]{Make sure to enter the correct
  conference title from your rights confirmation email}{June 03--05,
  2018}{Woodstock, NY}
\acmISBN{978-1-4503-XXXX-X/2018/06}

\usepackage{colortbl}  
\usepackage{xcolor}
\usepackage{multirow}
\usepackage[most]{tcolorbox}
\usepackage{pifont}
\usepackage{enumitem}
\usepackage{color}
\usepackage{subcaption}
\usepackage{graphicx}               

\NewDocumentCommand{\wei}
{ mO{} }{\textcolor{red}{\textsuperscript{\textit{Wei}}\textsf{\textbf{\small[#1]}}}}

\begin{document}

\title{Two Birds with One Stone: Skill-Augmented Generation via Unified Retrieval and Utilization}

\author{Zihan Luo}
\email{zihanluo@cqu.edu.cn}
\affiliation{%
  \institution{Chongqing University}
  \city{Chongqing}
  \country{China}
}

\author{Jiahui Kang}
\email{jiahuikang@stu.cqu.edu.cn}
\affiliation{%
  \institution{Chongqing University}
  \city{Chongqing}
  \country{China}
}

\author{Wei Zhou}
\email{weizhou2021@hust.edu.cn}
\affiliation{%
  \institution{Huazhong University of Science and Technology}
  \city{Wuhan}
  \country{China}}
  
\author{Qilin Fan}
\email{fanqilin@cqu.edu.cn}
\affiliation{%
  \institution{Chongqing University}
  \city{Chongqing}
  \country{China}
}

\author{Mingyan Li}
\email{limy2021@cqu.edu.cn}

\affiliation{%
  \institution{Chongqing University}
  \city{Chongqing}
  \country{China}
}

\author{Chao Chen}\authornote{Chao Chen is the corresponding author.}
\email{cschaochen@cqu.edu.cn}
\affiliation{%
  \institution{Chongqing University}
  \city{Chongqing}
  \country{China}
}
\renewcommand{\shortauthors}{Anonymous et al.}
\def\modelname{Skill-Uni}

\begin{abstract}
Skills, as reusable external knowledge, have demonstrated substantial potential to enhance the ability of \textit{Large Language Models} (LLMs) to solve complex tasks. However, existing skill-augmented generation methods typically rely on separate models for skill retrieval and skill utilization, resulting in disconnected optimization of these two processes. To address this gap, we propose \modelname{}, a unified reinforcement learning framework that jointly optimizes skill retrieval and utilization. Specifically, we adopt a generative skill retrieval paradigm that enables a single model to both learn how to retrieve relevant skills and leverage them for task solving. To effectively train this unified framework, we introduce a two-stage training strategy consisting of warm-up instruction tuning and GRPO, allowing the model to progressively improve its skill retrieval and utilization capabilities. Extensive experiments on five skill-related benchmarks demonstrate the effectiveness of \modelname{}, yielding average performance gains of 66.59\% and 34.82\% over the base Qwen3-4B and LLaMA3.1-8B models, respectively. Besides, \modelname{} is compatible with various skill retrieval engines, demonstrating its flexibility and generalizability. Additional in-depth analyses also provide insights into the effectiveness of \modelname{}. 
\end{abstract}


\maketitle

\input{introduction.tex}
\input{related_work.tex}
\input{preliminary.tex}
\input{method.tex}
\input{experiment.tex}
\input{conclusion.tex}

\bibliographystyle{ACM-Reference-Format}
\bibliography{sample-base}

\end{document}

%% file: introduction.tex
\section{Introduction}

\begin{figure}[t]
  \centering
  \includegraphics[width=0.95\linewidth]{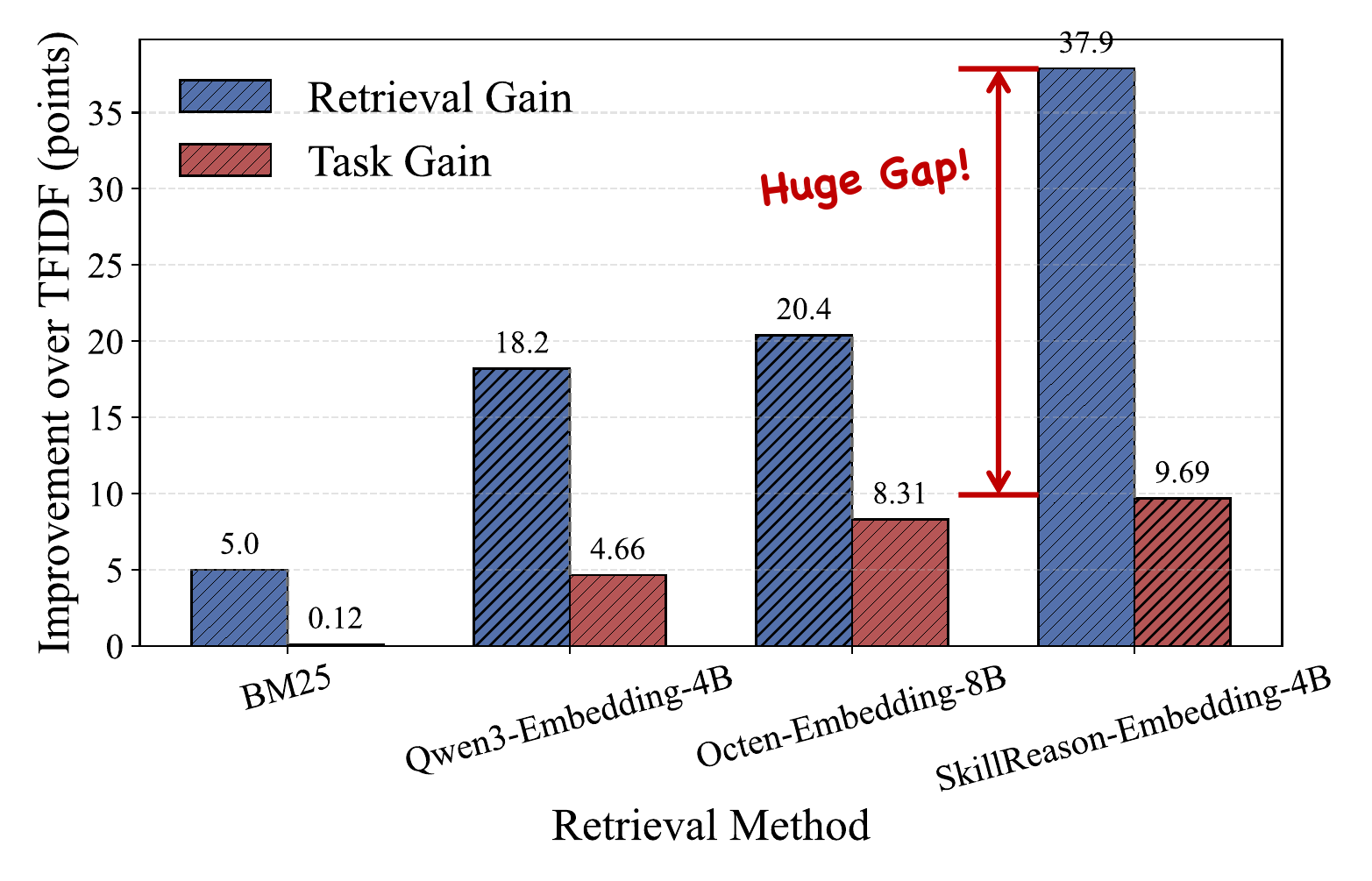}
  \vspace{-0.1cm}
  \caption{Retrieval and task accuracy gains relative to TF-IDF, measured as the accuracy differences between each retriever and TF-IDF in skill retrieval and downstream task completion, respectively. Although stronger retrievers achieve substantial improvements in retrieval accuracy, only part of these gains translate into downstream task performance.}
  \vspace{-0.2cm}
  \label{fig:motivation}
\end{figure}

As \textit{Large Language Models} (LLMs) continue to evolve from static question-answering systems into agents capable of autonomously planning and executing complex tasks, relying solely on knowledge encoded in model parameters is increasingly insufficient for real-world demands such as tool use, environment interaction, and long-horizon task execution. To extend the executable capabilities of LLM agents, recent work has increasingly organized reusable task-solving capabilities into modular “skills,” which describe their functionality, applicable conditions, and usage patterns in natural language and may further include instructions, scripts, and other auxiliary resources. Such skills can be retrieved, composed, and reused, and have become an important component of complex agent systems in domains such as embodied AI~\cite{skild}, software engineering~\cite{skillsbench}, and others.

As skill libraries continue to grow, efficiently identifying the skills most relevant to a given task has emerged as a key challenge. Existing work largely follows a retrieval-then-utilize paradigm~\cite{skillreason,skillret,srabench}, in which a small set of candidate skills is first retrieved from a skill library based on the user query and then provided to an LLM as external context for downstream task completion. Along this line, Cho et al. introduced SkillRet~\cite{skillret}, a large-scale benchmark for skill retrieval in LLM agents, and improved skill recall through dedicated retriever training. Building on this direction, Jiang et al. proposed SkillReason~\cite{skillreason} to address the difficulty of inferring latent skill needs from short and underspecified user requests. By incorporating reasoning-trace supervision and staged training, SkillReason integrates explicit reasoning into query representation and further improves dense skill retrieval.

Despite these advances, two important challenges still remain. First, \textit{retrieving the right skill does not guarantee using it effectively}. Unlike conventional retrieval-augmented generation~\cite{linearrag,lightrag}, where retrieved content is primarily factual text, skills often contain richer and more structured information, including instructions, scripts, tool interfaces, and auxiliary resources. Successful task completion therefore requires not only retrieving a relevant skill, but also applying it correctly in subsequent reasoning and execution. As shown in Figure~\ref{fig:motivation}, our empirical analysis on SRA-bench~\cite{srabench} with multiple retrievers further shows that improvements in retrieval accuracy translate only partially into downstream task gains, suggesting that skill utilization has become an important bottleneck. Second, \textit{skill retrieval and skill utilization are typically optimized in isolation}. Existing approaches often treat the retriever and the downstream LLM as separate components: the former is optimized for retrieval accuracy, while the latter passively consumes the retrieved content. However, skill retrieval and utilization are inherently coupled, as the value of a retrieved skill ultimately depends on its contribution to downstream task performance. Optimizing these two processes separately may therefore lead to a mismatch between retrieval objectives and actual task-solving needs.

\begin{figure}[t]
  \centering
  \includegraphics[width=0.98\linewidth]{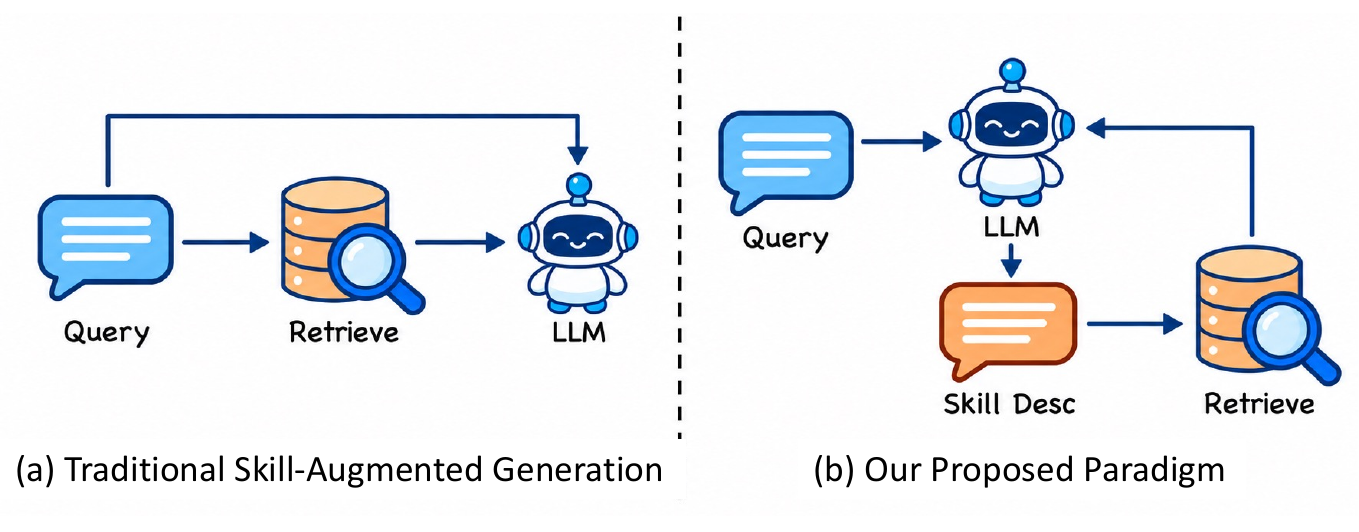}
  \vspace{-0.3cm}
  \caption{Overview of the traditional skill-augmented generation and our paradigm.}
  \vspace{-0.2cm}
  \label{fig:paragidm}
\end{figure}

To address these limitations, we propose \textbf{\modelname{}}, an end-to-end optimization framework that unifies skill retrieval and utilization. As illustrated in Figure~\ref{fig:paragidm}, rather than directly matching the original query against the skill library, \modelname{} adopts a generative skill retrieval paradigm, where the model dynamically generates descriptions of needed skills, retrieves relevant skills through an external retrieval engine, and integrates them into its ongoing generation process. This formulation provides a unified foundation for the model to jointly determine whether and how to retrieve, as well as how to utilize the retrieved skills, within a single generation trajectory. 
Specifically, to optimize these capabilities, we introduce a two-stage training strategy. We first perform warm-up instruction tuning on high-quality trajectories generated by a stronger teacher model, followed by outcome-based reinforcement learning that jointly optimizes retrieval decisions and the model's ability to leverage retrieved skills for task completion. By propagating task-level rewards through the entire generation trajectory, \modelname{} learns not only to retrieve useful skills but also to effectively apply them to solve downstream tasks. We further employ a token-masking strategy to stabilize training. Experiments on five skill-related benchmarks~\cite{srabench} demonstrate that \modelname{} achieves state-of-the-art performance in both task completion and skill retrieval, while exhibiting strong generalization and robustness.

Our contributions can be summarized as follows:
\begin{itemize}[leftmargin=*]
    \item We introduce \modelname{}, a unified framework for skill retrieval and utilization. Unlike conventional retrieval-then-utilize pipelines that treat retrieval and downstream reasoning as separate stages, \modelname{} formulates skill retrieval as a generative process and unifies whether to retrieve, what to retrieve, and how to utilize the retrieved skill within a single task formulation.
    \item We develop an effective training recipe for skill-augmented reasoning. The training procedure combines warm-up instruction tuning with reinforcement learning, enabling the model to first acquire the proposed interaction pattern and then jointly improve skill retrieval and utilization. We further introduce token masking to exclude externally retrieved skill content from the optimization objective, ensuring that learning focuses on model-generated retrieval and reasoning behaviors.
    \item Extensive experiments are conducted across multiple datasets, which demonstrate that \modelname{} consistently improves both Uni-retrieval quality and downstream task performance, while additional analyses highlight its generalization, robustness, and the importance of jointly optimizing skill retrieval and utilization.
\end{itemize}

%% file: related_work.tex
\section{Related Work}

\subsection{Agent Skills}

Recent advances in LLM agents have motivated the organization of reusable knowledge and task-solving capabilities into modular \emph{skills}. A representative example is Claude Agent Skills, where each skill is packaged with lightweight metadata, procedural instructions, and optional supporting
resources such as scripts, figures, and reference files. Skills are loaded on demand, allowing agents to acquire task-specific expertise without placing
the full skill library in the context window. Conceptually, skills are closely related to external tools~\cite{toolrec,toolace} and \textit{Retrieval-Augmented Generation} (RAG)~\cite{DBLP:conf/nips/LewisPPPKGKLYR020,DBLP:conf/www/DongAWL0S0CM25}, but serve a different role. Whereas tools primarily expose executable functions
and RAG typically provides factual or declarative knowledge, skills package reusable procedural knowledge that specifies how an agent should solve
a class of tasks. This modularity enables skills to be composed and reused across complex agentic settings, such as embodied AI~\cite{skild}, software engineering ~\cite{sweskillsbench}, and mathematical reasoning~\cite{skillsbench}. Benchmarks such as SRA-bench~\cite{srabench} and SkillsBench~\cite{skillsbench} have verified that proper skill integration can provide consistent improvements in downstream tasks given the external knowledge provided by skills. However, how to accurately locate and utilize relevant skills from a large-scale skill library continues to be a central bottleneck.

\subsection{Skill Retrieval}

With the rapid expansion of skill libraries, accurate skill retrieval has emerged as a rising research issue. Early skill-augmented systems typically rely on lexical retrieval methods such as BM25 and TF-IDF~\cite{srabench}. Although computationally efficient, these approaches primarily capture surface-level lexical overlap and may struggle when user requests express the required capability only implicitly.
Recent work has therefore focused on learning specialized retrievers for skill retrieval. SkillRet~\cite{skillret} introduces a large-scale benchmark and retrieval-oriented training data for skill selection, showing that task-specific fine-tuning substantially improves retrieval performance over off-the-shelf embedding models. SkillRouter~\cite{skillrouter} further adopts a retrieve-and-rerank pipeline and demonstrates that the full skill body can provide important signals beyond short metadata descriptions when distinguishing functionally similar skills. More recently, SkillReason~\cite{skillreason} targets implicit user requests by introducing reasoning supervision during training, enabling the retriever to better infer latent capability requirements while retaining efficient query-only retrieval at inference. However, all these works are still limited to skill retrieval, while largely overlooking skill utilization, the ability to interpret and apply the retrieved skill to complete the task.

\subsection{Skill Utilization}

Beyond retrieving relevant skills, effectively utilizing them is equally important, as skills often contain complex instructions, executable code, and other resources. Recent studies have explored different approaches to improving skill utilization. Skill0~\cite{skill0} introduces a training-time curriculum that progressively removes external skill guidance, encouraging models to internalize skill knowledge into their parameters through reinforcement learning. SelSkill~\cite{selskill} investigates whether a retrieved skill should be invoked or skipped, introducing a dual-granularity preference-learning framework to optimize skill invocation decisions.
Another line of research focuses on the dynamic evolution of skill libraries. SAGE~\cite{sage} introduces Sequential Rollout to accumulate reusable skills across related tasks and employs a skill-integrated reward to improve skill generation and utilization. Skill1~\cite{skill1} jointly optimizes skill selection, utilization, and distillation under a unified reinforcement learning framework, enabling the skill library to evolve through task experience. RASO~\cite{raso} leverages knowledge retrieved from existing skills to initialize and iteratively refine skills, incorporating execution feedback to guide subsequent updates.

%% file: preliminary.tex
\section{Preliminary}
\subsection{Skill-Augmented Generation}

As reusable modules of experience and knowledge, skills are widely employed to assist AI agents in tackling complex tasks. Formally, let $\mathcal{C}=\{\mathcal{S}_1, \mathcal{S}_2, ..., \mathcal{S}_N\}$ denote a skill library that contains $N$ different skills, and each skill $\mathcal{S}_i$ can be represented as: $\mathcal{S}_i = (\mathcal{N}_i, \mathcal{D}_i, \mathcal{I}_i, \mathcal{P}_i)$, where $\mathcal{N}_i$ denotes the skill name, $\mathcal{D}_i$ represents a concise description, $\mathcal{I}_i$ specifies task-relevant instructions, and $\mathcal{P}_i$ encompasses additional payloads provided within the skill, such as scripts, references, and digital assets. 

To efficiently navigate large-scale skill libraries, skill-augmented agents typically adopt a progressive disclosure paradigm~\cite{srabench}. Given a task query $q$, a retriever $\mathcal{R}$ first identifies relevant skill candidates using lightweight metadata, such as skill names $\mathcal{N}_i$ and descriptions $\mathcal{D}_i$. Formally, the retriever $\mathcal{R}$ selects a small relevant candidate set $\mathcal{C}_r$ from the skill library $\mathcal{C}$:
\begin{equation}
    \mathcal{C}_r = \mathcal{R}(q, \{(\mathcal{N}_i, \mathcal{D}_i)\}_{i=1}^N),
\end{equation}
where $\mathcal{C}_r \subset \mathcal{C}$ and $|\mathcal{C}_r| = k \ll N$. Only after $\mathcal{C}_r$ is selected does the agent load the full content of these $k$ skills into its context window, preventing context length explosion while ensuring precise capability provision:
\begin{equation}
    \mathcal{A}=\mathcal{M}(q,\mathcal{C}_r),
\end{equation}
where $\mathcal{M}$ denotes the AI agent empowered by a large language model, and $\mathcal{A}$ represents the final answer.

\subsection{Group Relative Policy Optimization}

Instead of learning a critic model to estimate the rewards as in \textit{Proximal Policy Optimization} (PPO)~\cite{ppo}, \textit{Group Relative Policy Optimization} (GRPO)~\cite{deepseekmath} is a reinforcement learning algorithm that estimates the rewards dynamically by sampling a group of candidate responses for each query and normalizing their rewards relative to the group. Formally, given a task query $q$, the current policy generates a group of $G$ candidate outputs $\{o_1, o_2, \dots, o_G\}$ with rewards $\{r_1, r_2, \dots, r_G\}$. To quantify the relative quality of each output within the group, the advantage $A_i$ for $o_i$ is computed via standardizing the group rewards:
\begin{equation}
A_i = \frac{r_i - \text{mean}(\{r_j\}_{j=1}^G)}{\text{std}(\{r_j\}_{j=1}^G)},
\end{equation}
and the policy model $\pi_\theta$ with parameters $\theta$ is then optimized by maximizing the following PPO-style clipped objective:
\begin{equation}
\begin{aligned}
\mathcal{J}_{\text{GRPO}}(\theta) = \mathbb{E}_{q, \{o_i\}} \left[ \frac{1}{G} \sum_{i=1}^{G} \min \left( \phi_i A_i, \right. \right. \\
\left. \vphantom{\frac{1}{G}} \left. \mathbf{clip}(\phi_i, 1 - \epsilon, 1 + \epsilon) A_i \right) - \beta D_{\text{KL}}(\pi_\theta \parallel \pi_{\text{ref}}) \right],
\end{aligned}
\end{equation}
where $\phi_i = \frac{\pi_\theta(o_{i} \mid q)}{\pi_{\text{ref}}(o_{i} \mid q)}$ represents the importance sampling ratio, $\epsilon$ is the clipping hyperparameter, and $\beta$ controls the strength of the KL penalty to prevent drift from the old reference policy $\pi_{\text{ref}}$.

%% file: method.tex
\section{Method}

\definecolor{tagRecColor}{RGB}{0, 128, 128}     
\definecolor{tagSkillColor}{RGB}{180, 50, 50}   
\definecolor{tagAnsColor}{RGB}{30, 100, 200}    

\newcommand{\rectag}[1]{\textcolor{tagRecColor}{\textbf{\texttt{#1}}}}
\newcommand{\skilltag}[1]{\textcolor{tagSkillColor}{\textbf{\texttt{#1}}}}
\newcommand{\anstag}[1]{\textcolor{tagAnsColor}{\textbf{\texttt{#1}}}}

\begin{figure*}[t]
  \centering
  \includegraphics[width=0.9\textwidth]{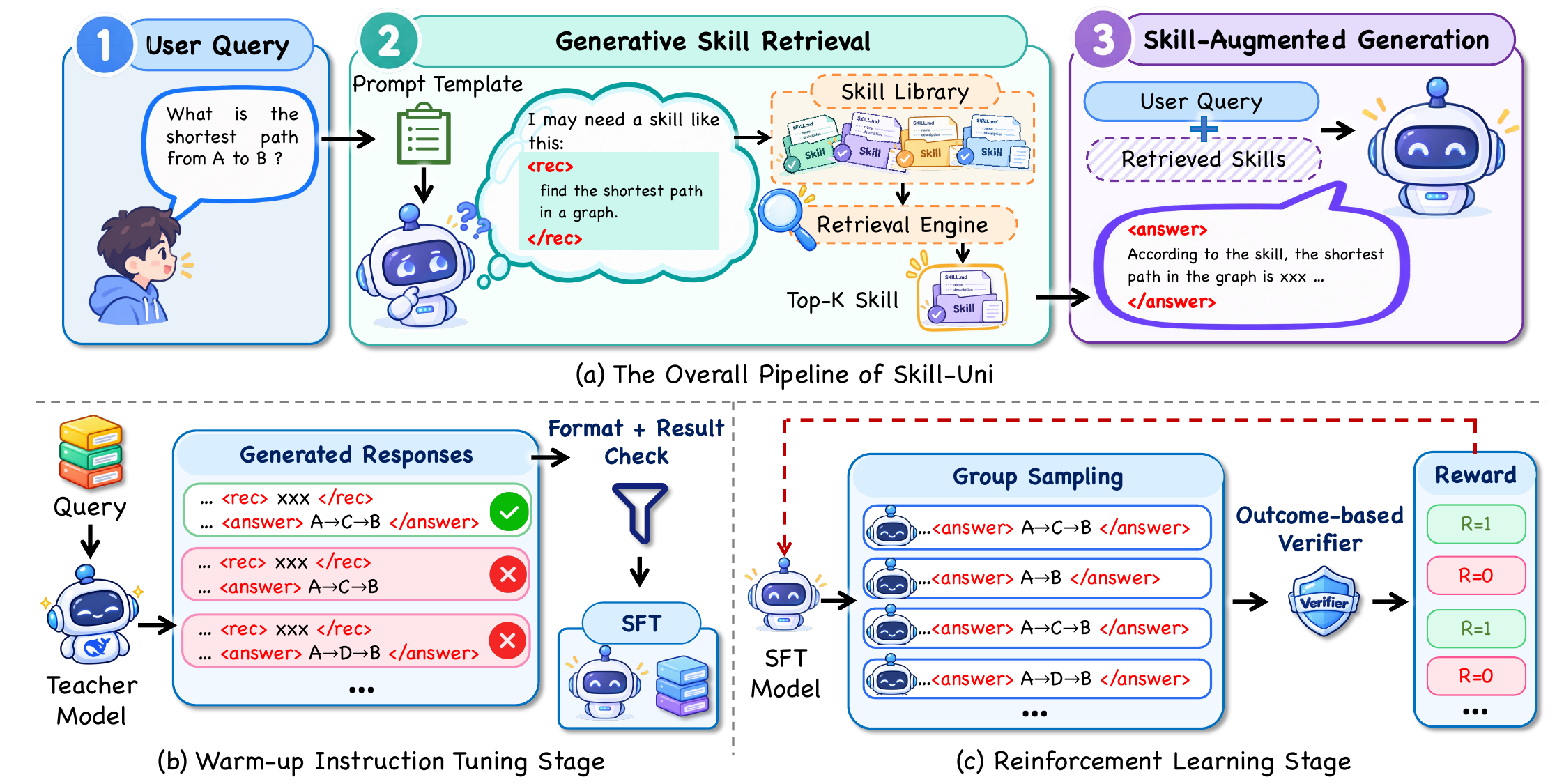}
  \caption{The overall framework of \modelname{}. (a) Given a query, \modelname{} is first prompted to describe its desired skills, which will be further used for skill retrieval, and then generate the final response with both the query and the retrieved skills. The whole training process consists of a warm-up instruction tuning stage (b) and a reinforcement learning stage (c).}
  \label{fig:framework}
\end{figure*}

\subsection{Overview}
In this paper, we introduce \modelname{}, a unified framework that jointly improves skill retrieval and utilization with a single model. As illustrated in Figure~\ref{fig:framework}, \modelname{} is first prompted to perform generative skill retrieval, where it generates a brief description of the skill needed for the query. The description is then used to retrieve the most relevant skill from the skill library, which guides the generation of the final response.
To equip the model with this capability, we adopt a two-stage training strategy. We first perform warm-up instruction tuning using high-quality teacher-generated trajectories filtered by format and answer correctness. The model is then refined with reinforcement learning using only outcome-based rewards, enabling \modelname{} to explore various skills and improve the capability to retrieve and use skills jointly through an end-to-end optimization manner.

\subsection{Unified Task Formulation}
Inspired by the paradigm in Search-R1~\cite{search-r1}, we treat the retrieval engine as an external tool that can be invoked through structured model outputs. As the prompt template shown below, rather than directly retrieving skills via the original query, \modelname{} is instructed to express the desired skill description in a designated \rectag{<rec>}...\rectag{</rec>} field. This generated description is subsequently passed to the retrieval engine, which will return the full content of the top-$K$ relevant skills in \skilltag{<skill>}...\skilltag{</skill>}. \modelname{} can then continue reasoning with the retrieved skill information and produce its final response in \anstag{<answer>}...\anstag{</answer>}. Note that, \modelname{} is agnostic to the underlying retrieval engine $\mathcal{R}$, and if no external skill is needed, the model can bypass retrieval altogether and directly generate an answer.  

\begin{tcolorbox}[
    colback=gray!5,             
    colframe=gray!60,            
    arc=2mm,                    
    boxrule=0.8pt,              
    left=5pt, right=5pt, top=4pt, bottom=4pt, 
    fontupper=\small\sffamily   
]
To answer the given question, you can use a skill retrieval engine by \rectag{<rec>} your desired skill's description \rectag{</rec>}, and it will return the top-1 related skill information between \skilltag{<skill>} and \skilltag{</skill>}, including the skill's description and content. You can retrieve as many times as you want. If you find no further external skill needed, you can directly answer the question inside \anstag{<answer>} and \anstag{</answer>}. \textbf{Question:} 
\end{tcolorbox}

This formulation offers two main advantages: 1) It places skill retrieval and skill utilization within the same model trajectory, allowing them to be jointly optimized rather than treated as isolated components. Because what should be retrieved depends on how the retrieved skill will ultimately be used, this unified formulation encourages better coordination between the two processes; 2) Generative skill retrieval can naturally bridge the semantic gap between user queries and skill descriptions by leveraging the superior reasoning capabilities of LLMs. As it is independent of the retrieval engine, \modelname{} can augment a wide range of existing retrievers $\mathcal{R}$ like BM25 and Qwen3-Embedding, improving retrieval effectiveness without requiring modifications to their core mechanisms.

\subsection{Warm-start Instruction Tuning}
Different from the cold-start reinforcement learning settings in existing work~\cite{search-r1,r1}, we empirically find that direct reinforcement learning optimization is less effective in our setting. In particular, when the task requires relatively complex reasoning and the available training data are limited, the base model often fails to produce sufficiently informative learning signals during RL, resulting in weak or unstable optimization progress. Therefore, during the training of \modelname{}, we first introduce a warm-up instruction tuning stage to improve the model’s basic instruction-following capability under our task formulation. Specifically, we use DeepSeek-V4-Pro\footnote{https://huggingface.co/deepseek-ai/DeepSeek-V4-Pro-0813} as the teacher model and prompt it to generate responses following the predefined template, after which the generated samples are filtered based on both answer correctness and format validity, where the latter requires all structured fields to be properly closed and a valid \anstag{<answer>}...\anstag{</answer>} field to be present.

Given the capability gap between the teacher and student models, we first evaluate the student model on each query and assign a label according to whether it can solve the query correctly without external skills. For queries that the student fails to answer correctly, DeepSeek-V4-Pro is instructed to generate a retrieval request within the \rectag{<rec>}...\rectag{</rec>} field, thereby providing supervision for skill retrieval. In contrast, for queries that the student can already solve directly, the teacher is instructed to produce the final answer in \anstag{<answer>}...\anstag{</answer>} without invoking skill retrieval. This capability-aware data construction strategy aligns the supervision signal with the student model’s actual need for external skills, rather than forcing retrieval uniformly across all samples.

After constructing the filtered training set, we adopt conventional \textit{Supervised Fine-tuning} (SFT) to optimize the model. Specifically, the objective can formulated as:
\begin{equation}
    \mathcal{L}_{\mathrm{SFT}}=-\frac{1}{N}\sum_{i=1}^{N}\sum_{t=1}^T\mathbb{I}(o_{i,t})log\text{P}(o_{i,t}|x_{i},o_{i,<t},\mathcal{R};\theta)
\end{equation}
where $N$ denotes the number of training data, and $x_{i}$ is the input context with the proposed prompt template.
Importantly, $\mathbb{I}(o_{i,t})$ is an indicator function that determines whether the current token $o_{i,t}$ is generated by LLMs. During SFT, we mask the retrieved tokens enclosed by \skilltag{<skill>}...\skilltag{</skill>} so the model can focus on learning to use the skill retrieval engine and apply the skill.

\subsection{Reinforcement Learning with Skill Retrieval}

Building on the model $\theta_{\mathrm{SFT}}$ obtained from the above warm-up instruction tuning, we further apply reinforcement learning to strengthen its skill retrieval and skill utilization capabilities. Specifically, given a skill retrieval engine $\mathcal{R}$, we adopt the GRPO training strategy~\cite{deepseekmath} and use an outcome-based verifier as the only source of reward:
\begin{equation}
r_i =
\begin{cases}
1, & \text{if $o_i$ is correct}, \\
0, & \text{otherwise}.
\end{cases}
\end{equation}
It is worth noting that we do not introduce any explicit reward for retrieving the gold skill, as such supervision is often difficult to obtain at scale and may overly constrain the model to a predefined retrieval target, potentially limiting its ability to explore alternative skills that are also useful for solving the task. Same with the warm-up instruction tuning stage, we employ a token mask strategy during GRPO as well to exclude retrieved tokens in the \skilltag{<skill>} field, avoiding potential distribution shift during training, and the training objective can be formulated as:
\begin{equation}
\begin{aligned}
\mathcal{J}_{\mathrm{GRPO}}(\theta)
=
\mathbb{E}_{q,\{o_i\}_{i=1}^{G}}
\Bigg[
\frac{1}{G}\sum_{i=1}^{G}
\frac{1}{\sum_t \mathbb{I}(o_{i,t})}
\sum_t \mathbb{I}(o_{i,t})
\min\Big(
\phi_{i,t}A_{i,t}, \\
\mathbf{clip}\big(\phi_{i,t},1-\epsilon,1+\epsilon\big)A_{i,t}
\Big)
-
\beta D_{\mathrm{KL}}
\Bigg],
\end{aligned}
\label{eq:grpo}
\end{equation}
where $A_{i,t}$ is the advantage computed by the relative rewards within the group, and $\phi_{i,t}$ is the importance sampling ratio:
\begin{equation}
\phi_{i,t}
=
\frac{
\pi_\theta(o_{i,t}|x_i, o_{i,<t},\mathcal{R})
}{
\pi_{\mathrm{\theta_{SFT}}}(o_{i,t}|x_i, o_{i,<t},\mathcal{R})
}.
\end{equation}

%% file: experiment.tex
\section{Experiments}

\begin{table*}[t]
\centering
\renewcommand{\arraystretch}{1.02}
\caption{Performance comparison on multiple datasets. The base model is Qwen3-4B. 
Task accuracy (\%) is reported, and the best performance is \textbf{bold}. 
Improvements in the Average column are measured against the Base model.}
\label{tab:main}
\resizebox{0.8\textwidth}{!}{%
\begin{tabular}{lccccc>{\columncolor[RGB]{245,244,252}}c}
\toprule
\textbf{Method} 
& \textbf{TheoremQA} 
& \textbf{LogicBench} 
& \textbf{CHAMP} 
& \textbf{MedCalc-Bench} 
& \textbf{BigCodeBench} 
& \textbf{Average} \\
\midrule

Base (Qwen3-4B)
& 49.66 & 78.95 & 51.11 & 29.55 & 32.46 
& 44.84 $(\Delta_{\mathrm{base}})$ \\

Oracle Skill 
& 71.81 & 86.84 & 68.89 & 55.91 & 34.21 
& 57.43 \textcolor{green!50!black}{$(\uparrow\,28.08\%)$} \\

\midrule
\rowcolor[RGB]{234,238,234}
\multicolumn{7}{l}{\textit{Lexical Matching Methods}} \\

BM25   
& 67.11 & 67.11 & 71.11 & 31.36 & 30.26 
& 46.85 \textcolor{green!50!black}{$(\uparrow\,4.48\%)$} \\

TF-IDF 
& 65.10 & 67.11 & 48.89 & 35.91 & 31.14 
& 46.73 \textcolor{green!50!black}{$(\uparrow\,4.21\%)$} \\

\midrule
\rowcolor[RGB]{234,238,234}
\multicolumn{7}{l}{\textit{Dense Embedding Retrieval Methods}} \\

Qwen3-Embedding-4B
& 64.43 & 68.42 & 68.89 & 48.18 & 31.14
& 51.39 \textcolor{green!50!black}{$(\uparrow\,14.61\%)$} \\

Octen-Embedding-8B
& 67.11 & 79.61 & 66.67 & 55.00 & 28.51
& 55.04 \textcolor{green!50!black}{$(\uparrow\,22.75\%)$} \\

SkillRet-Embedding-0.6B
& 63.09 & 71.71 & 66.67 & 54.55 & 29.39
& 52.90 \textcolor{green!50!black}{$(\uparrow\,17.98\%)$} \\

SkillRouter-Embedding-0.6B
& 63.09 & 72.37 & 71.11 & 48.64 & 29.82
& 51.76 \textcolor{green!50!black}{$(\uparrow\,15.43\%)$} \\

SkillReason-Embedding-4B
& 69.80 & 77.63 & \textbf{75.56} & 54.55 & 31.58
& 56.42 \textcolor{green!50!black}{$(\uparrow\,25.82\%)$} \\

\midrule
\rowcolor[RGB]{234,238,234}
\multicolumn{7}{l}{\textit{LLM-based Retrieval Methods}} \\

Qwen3-4B
& 63.09 & 71.71 & 68.89 & 51.82 & 32.02
& 53.02 \textcolor{green!50!black}{$(\uparrow\,18.24\%)$} \\

DeepSeek-V4-Pro
& 71.81 & 75.00 & 68.89 & 55.91 & \textbf{34.21}
& 57.05 \textcolor{green!50!black}{$(\uparrow\,27.23\%)$} \\

\midrule

\modelname{}
& \textbf{71.81}
& \textbf{85.53}
& 62.22
& \textbf{68.64}
& 28.07
& \textbf{60.45} \textcolor{green!50!black}{$(\uparrow\,34.81\%)$} \\

\bottomrule
\end{tabular}%
}
\end{table*}

\begin{table*}[t]
\centering
\renewcommand{\arraystretch}{1.02}
\caption{Performance comparison on multiple datasets. The base model is LLaMA3.1-8B. 
Task accuracy (\%) is reported, and the best performance is \textbf{bold}. 
Improvements in the Average column are measured against the Base model.}
\label{tab:main_llama}
\resizebox{0.8\textwidth}{!}{%
\begin{tabular}{lccccc>{\columncolor[RGB]{245,244,252}}c}
\toprule
\textbf{Method} 
& \textbf{TheoremQA} 
& \textbf{LogicBench} 
& \textbf{CHAMP} 
& \textbf{MedCalc-Bench} 
& \textbf{BigCodeBench} 
& \textbf{Average} \\
\midrule

Base (LLaMA3.1-8B)
& 31.59 & 53.42 & 21.52 & 26.82 & 29.21 
& 32.51 $(\Delta_{\mathrm{base}})$ \\

Oracle Skill 
& 47.26 & 68.03 & 42.15 & 60.82 & 29.47 
& 49.55 \textcolor{green!50!black}{$(\uparrow\,52.41\%)$} \\

\midrule
\rowcolor[RGB]{234,238,234}
\multicolumn{7}{l}{\textit{Lexical Matching Methods}} \\

BM25   
& 39.76 & 57.24 & 23.77 & 35.91 & 28.95 
& 37.12 \textcolor{green!50!black}{$(\uparrow\,14.18\%)$} \\

TF-IDF 
& 35.74 & 49.61 & 24.66 & 38.45 & 29.39 
& 35.57 \textcolor{green!50!black}{$(\uparrow\,9.51\%)$} \\

\midrule
\rowcolor[RGB]{234,238,234}
\multicolumn{7}{l}{\textit{Dense Embedding Retrieval Methods}} \\

Qwen3-Embedding-4B
& 43.62 & 57.24 & 52.73 & 15.56 & \textbf{34.65}
& 44.58 \textcolor{green!50!black}{$(\uparrow\,37.13\%)$} \\

Octen-Embedding-8B
& 46.31 & 62.50 & 58.64 & 22.22 & 27.19
& 45.97 \textcolor{green!50!black}{$(\uparrow\,41.40\%)$} \\

SkillRet-Embedding-0.6B
& 40.94 & 57.89 & 54.09 & 22.22 & 31.58
& 44.08 \textcolor{green!50!black}{$(\uparrow\,35.59\%)$} \\

SkillRouter-Embedding-0.6B
& 41.61 & 56.58 & 51.36 & 35.56 & 25.44
& 42.19 \textcolor{green!50!black}{$(\uparrow\,29.78\%)$} \\

SkillReason-Embedding-4B
& 44.30 & 70.39 & 58.18 & \textbf{42.22} & 32.89
& 49.75 \textcolor{green!50!black}{$(\uparrow\,53.03\%)$} \\

\midrule
\rowcolor[RGB]{234,238,234}
\multicolumn{7}{l}{\textit{LLM-based Retrieval Methods}} \\

Qwen3-4B
& 42.95 & 59.87 & 53.18 & 20.00 & 24.56
& 42.44 \textcolor{green!50!black}{$(\uparrow\,30.54\%)$} \\

DeepSeek-V4-Pro
& 46.98 & 64.47 & 56.36 & 35.56 & 31.58
& 47.86 \textcolor{green!50!black}{$(\uparrow\,47.22\%)$} \\

\midrule

\modelname{}
& \textbf{55.03}
& \textbf{84.21}
& \textbf{62.73}
& 35.56
& 28.95
& \textbf{54.16} \textcolor{green!50!black}{$(\uparrow\,66.59\%)$} \\

\bottomrule
\end{tabular}%
}
\end{table*}

\subsection{Experimental Setup}

\textbf{Dataset.}
Our data is primarily derived from SRA-bench~\cite{srabench}, which provides ground-truth answers alongside golden skill annotations for each query. Specifically, we select five subsets from SRA-bench, including TheoremQA~\cite{theoremqa}, LogicBench~\cite{logicbench}, CHAMP~\cite{champ}, MedCalc-Bench~\cite{medcalcbench}, and BigCodeBench~\cite{bigcodebench}. For each subset, we perform a random split by allocating 80\% of the queries for training and reserving the remaining 20\% for test. To further evaluate the generalization capabilities of \modelname{}, we additionally conduct experiments on MBPP+~\cite{mbpp+} and LiveCodeBench~\cite{livecodebench}.

\textbf{Baselines.}
Following the settings of prior work~\cite{srabench,skillreason}, we primarily focus on baselines under the retrieve-then-utilize paradigm, where the task models are kept frozen. According to their retrieval mechanisms, these baselines can be categorized into the following three distinct categories:
\begin{itemize}[leftmargin=0.5cm]
    \item \textit{Lexical matching methods}: This category relies on surface-level word overlaps and term frequency statistics. Specifically, we include two representative methods namely BM25 and TF-IDF~\cite{tfidf} as baselines.
    \item \textit{Dense embedding retrieval methods}: To capture semantic intent beyond surface lexical matches, these methods encode queries and skills into a shared dense vector space. We consider both general-purpose retrievers like Qwen3-Embedding-4B\footnote{https://huggingface.co/Qwen/Qwen3-Embedding-4B} and Octen-Embedding-8B\footnote{https://huggingface.co/Octen/Octen-Embedding-8B}, and skill-specific embedding models, including SkillRet-Embedding-0.6B~\cite{skillret}, SkillRouter-Embedding-0.6B~\cite{skillrouter}, SkillReason-Embedding-4B~\cite{skillreason}.
    \item \textit{LLM-based retrieval methods}: These approaches leverage the reasoning capabilities of LLMs to select the most relevant skills from a top-50 candidate pool initially retrieved by BM25. Specifically, Qwen3-4B\footnote{https://huggingface.co/Qwen/Qwen3-4B} and DeepSeek-V4-Pro are employed as the retrieval models.
\end{itemize}

\textbf{Implementation details.}
All models are trained and evaluated on a single Linux server equipped with two Intel Xeon Platinum 8350C processors (2.60 GHz), and two NVIDIA A100 80 GB GPUs with CUDA 12.8. We employ GRPO~\cite{deepseekmath} as our reinforcement learning strategy, and both the SFT and GRPO are implemented with the verl framework~\cite{verl} and vLLM~\cite{vllm}. Specifically, we fine-tune Qwen3-4B and LLaMA3.1-8B\footnote{https://huggingface.co/meta-llama/Llama-3.1-8B} with LoRA~\cite{lora} using rank 16 and $\alpha=32$, applied to all linear layers. The AdamW optimizer is employed with a learning rate of $1\times10^{-4}$, cosine learning-rate decay, a warmup ratio of 0.03, and weight decay of 0.01. The group size $G$ for GRPO is set to 8, and we train 1 epoch and 2 epochs for SFT and GRPO, respectively. SkillReason-Embedding-4B~\cite{skillreason} is set as the default retriever, and retrieves the top-1 skill during training. Alternative retrievers will also be discussed later.

\subsection{Main Results}
We compare \modelname{} with multiple baselines across the five subsets of SRA-bench, and report their task completion accuracy. For a fair comparison, all baselines employ the same base model for task completion. We also report the performance of Oracle, where the model is provided with a human-annotated golden skill. Specifically, Qwen3-4B and LLaMA3.1-8B are deployed as the base models, respectively. As presented in Table~\ref{tab:main} and Table~\ref{tab:main_llama}, the following key observations can be drawn:
\begin{itemize}[leftmargin=*]
    \item Incorporating skill retrieval leads to clear performance improvements across all datasets and base models. Specifically, both lexical matching and dense embedding retrieval methods outperform direct generation. This finding aligns with prior work~\cite{srabench}, confirming that external modular skills can effectively empower models to tackle complex tasks.

    \item Dense embedding and LLM-based retrieval methods generally surpass lexical matching baselines. For example, SkillReason-Embedding-4B achieves 56.42\% average accuracy in Table~\ref{tab:main}, largely outperforming BM25 and TF-IDF. This superiority stems from their stronger semantic comprehension, which enables more accurate retrieval of target skills and highlights the necessity of developing advanced skill retrieval strategies.

    \item With two different base models, \modelname{} achieves superior performance across multiple datasets and establishes a new state-of-the-art average accuracy. Remarkably, \modelname{} outperforms even the Oracle baseline by a noticeable margin in average accuracy (60.45\% vs. 57.43\% and 54.16\% vs. 49.55\%).
    This substantial gain indicates that the benefits of \modelname{} derive not only from precise skill retrieval, but also from its superior skill utilization.
\end{itemize}

\subsection{Ablation Study}

To systematically investigate the individual contributions of key components in \modelname{}, we design five ablation variants: \textit{(i) w/o Mask}: removing the token mask strategy during SFT and GRPO;
\textit{(ii) w/o Warm-Start SFT}: removing the supervised fine-tuning stage prior to RL; 
\textit{(iii) w/o GRPO}: eliminating the reinforcement learning stage and relying solely on SFT; 
\textit{(iv) w/o Generative Retrieval}: utilizing the raw input query directly for skill retrieval instead of generating a \rectag{<rec>} field; and 
\textit{(v) Frozen Task Execution}: employing a frozen base model instead of \modelname{} for downstream task execution, without the use of the \anstag{<answer>} field.

As reported in Table~\ref{tab:ablation}, after removing the mask strategy during training, the variant only achieves 46.47\% accuracy, far behind the full version. Besides, removing either the SFT stage or the GRPO stage leads to a substantial degradation in overall performance. This degradation highlights the complementary nature of the two stages: the SFT stage establishes formatting and reasoning priors for generating \rectag{<rec>} and \anstag{<answer>} fields, while GRPO drives strategy refinement through policy gradients. Furthermore, replacing the generative \rectag{<rec>} field with the raw query or resorting to a frozen task model also incurs notable performance drops. These findings highlight the importance of both accurate skill retrieval and robust skill utilization for successful task completion.

\begin{table}[t]
\centering
\caption{Ablation study of \modelname{}. Average task accuracy (\%) on five datasets is reported. The base model is Qwen3-4B.}
\vspace{-0.1cm}
\label{tab:ablation}
\renewcommand{\arraystretch}{1.05}
\begin{small}
\resizebox{0.85\linewidth}{!}{
    \begin{tabular}{c | c c c c c|c}
        \toprule
        \multicolumn{6}{c|}{\textbf{Model Variants}} & \multirow{2}{*}{\textbf{Accuracy}}\\
        \cmidrule{1-6}
        ID & mask & SFT & GRPO & \rectag{<rec>} & \anstag{<answer>} 
        \\
        \midrule
        1       & \ding{55}    & \ding{51} & \ding{51}   & \ding{51}   & \ding{51} & 46.47 \\
        2       & \ding{51} & \ding{55}   & \ding{51}   & \ding{51}   & \ding{51} & 51.89 \\
        3       & \ding{51} & \ding{51} & \ding{55}   & \ding{51}  & \ding{51} & 51.64 \\
        4       & \ding{51} & \ding{51} & \ding{51} & \ding{55}  & \ding{51} & 59.57 \\
        5       & \ding{51} & \ding{51} & \ding{51} & \ding{51}  & \ding{55}   & 57.93 \\ \midrule
        \modelname{} & \ding{51} & \ding{51} & \ding{51} & \ding{51} & \ding{51} & \textbf{60.45} \\
        \bottomrule
    \end{tabular}}
\end{small}
\end{table}

\subsection{Stratified Performance Analysis}
\begin{figure}[t]
  \centering
  \includegraphics[width=0.95\columnwidth]{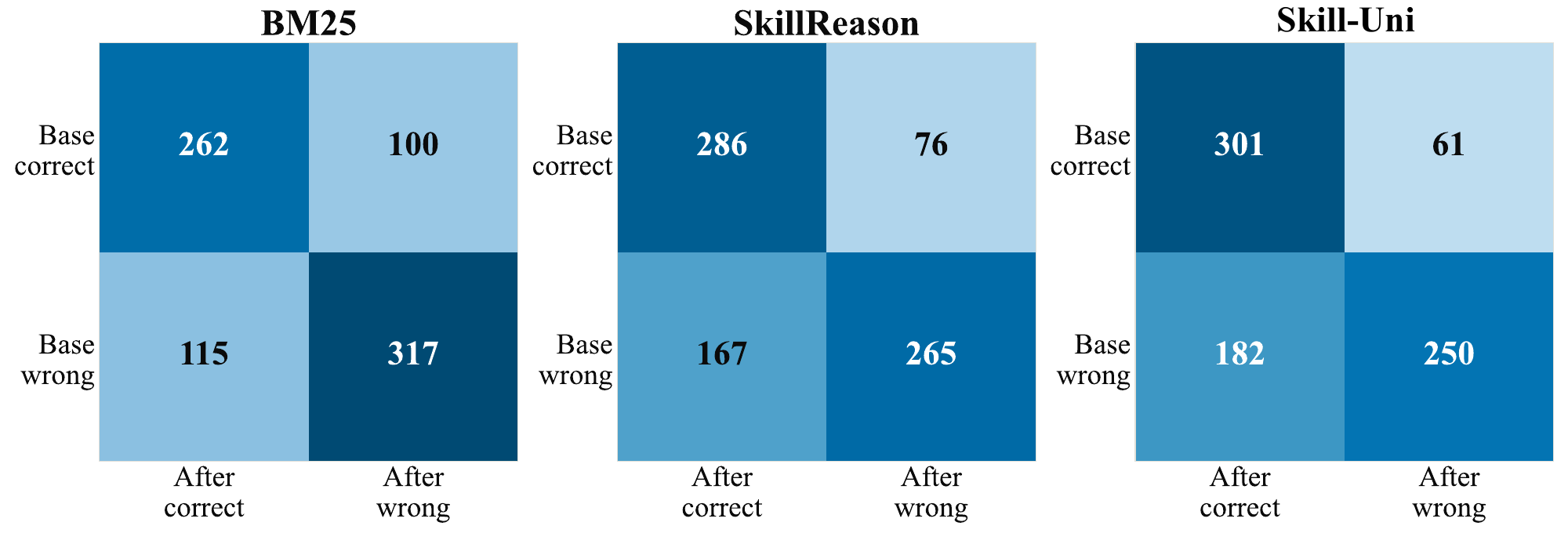}
  \caption{The stratified performance based on query correctness before and after applying a given method.}
  \label{fig:confusion}
\end{figure}

In this section, we conduct a fine-grained stratified analysis to gain deeper insight into model behavior. Specifically, based on query correctness before and after applying a given method, we partition all queries into four distinct groups: \textit{(i)} Incorrect $\rightarrow$ Correct, \textit{(ii)} Correct $\rightarrow$ Correct, \textit{(iii)} Correct $\rightarrow$ Incorrect, and \textit{(iv)} Incorrect $\rightarrow$ Incorrect. Figure~\ref{fig:confusion} illustrates the distribution of query counts across these categories for BM25, SkillReason, and \modelname{}, respectively. The base model is Qwen3-4B.

The results reveal two clear advantages of \modelname{}. First, it converts the largest number of originally incorrect queries into correct (182), indicating a stronger ability to recover from failures by retrieving and applying useful skills. Second, \modelname{} produces the fewest Correct $\rightarrow$ Incorrect cases (61), suggesting that it is less likely to disrupt queries that the model can already solve correctly. This distinction is important because retrieval does not always improve performance: irrelevant or unnecessary skills may introduce distracting information and degrade otherwise correct predictions. We attribute the stronger performance of \modelname{} to the combination of more accurate skill retrieval and more effective skill utilization, which together help the model introduce external skills when they are beneficial while reducing the effects of irrelevant skill context.

\subsection{In-depth Analysis}
\begin{figure}[t]
  \centering
  \begin{subfigure}[b]{0.48\columnwidth}
    \includegraphics[width=\textwidth]{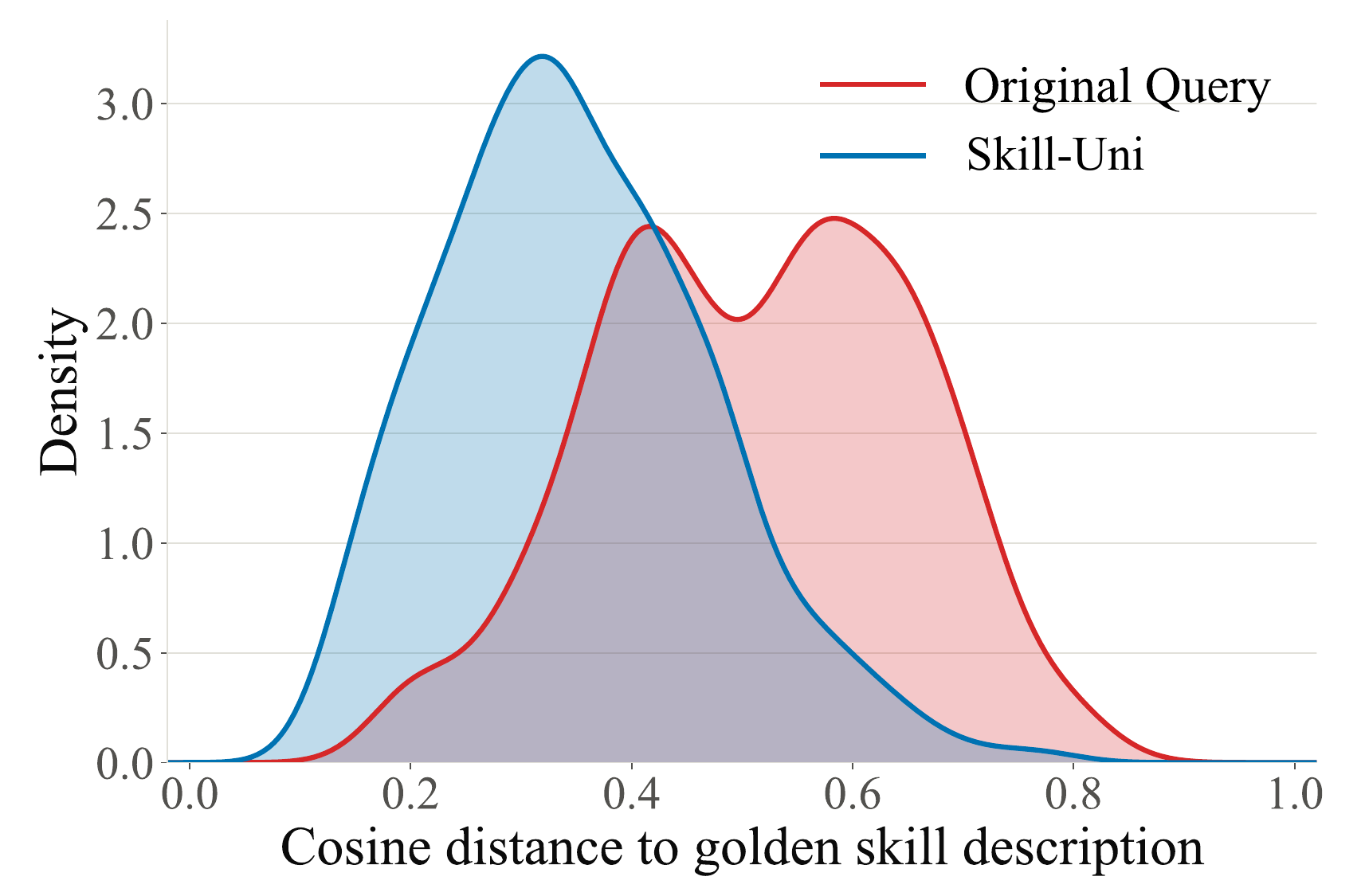}
    \caption{The distance distribution of different retrieval mechanisms}
    \label{fig:density}
  \end{subfigure}
  \hfill
  \begin{subfigure}[b]{0.48\columnwidth}
    \includegraphics[width=\textwidth]{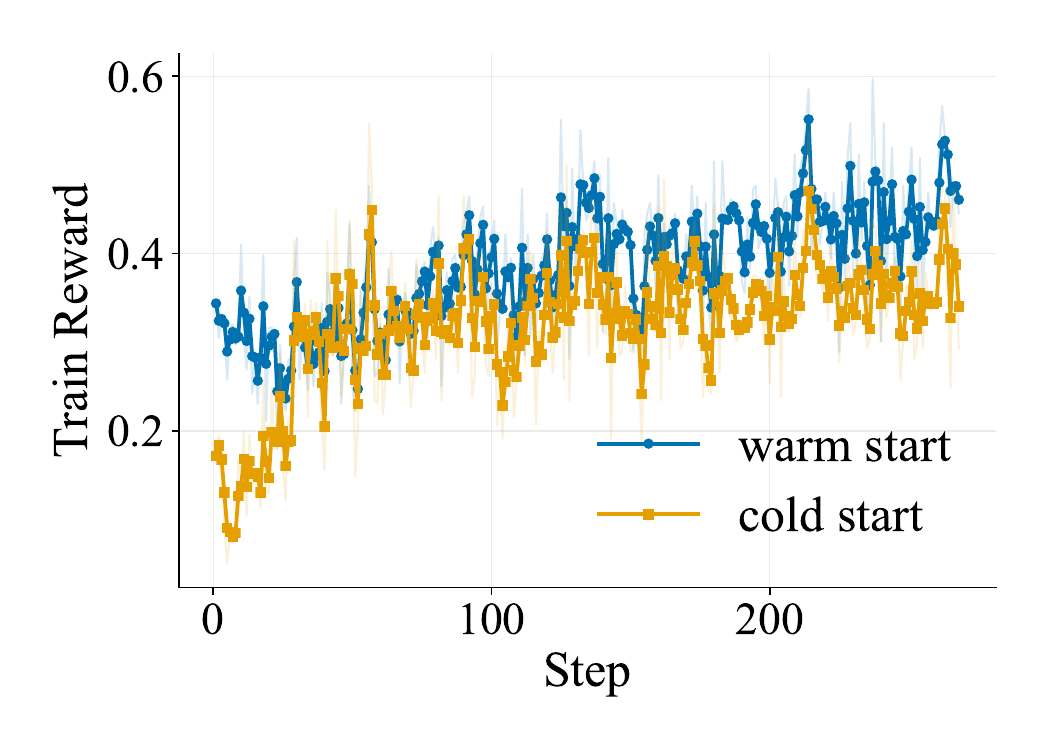}
    \caption{RL training curves w/ and w/o warm-start instruction tuning}
    \label{fig:rl_nowarm}
  \end{subfigure}
  \vspace{-0.1cm}
  \caption{The results of in-depth analysis on (a) the effect of generative skill retrieval and (b) the effect of warm start. The base model is Qwen3-4B.}
  \label{fig:rl_curve}
\end{figure}

In this part, we further provide a series of in-depth model analyses from the following two perspectives:

\textbf{The effect of generative skill retrieval}. 
Unlike existing baselines, \modelname{} introduces a generative retrieval paradigm that actively generates desired skill descriptions within a \rectag{<rec>}...\rectag{</rec>} field to drive skill retrieval. To verify its effectiveness, we compare the embedding distances of original queries versus generated \rectag{<rec>}...\rectag{</rec>} fields against golden skill descriptions, leveraging Qwen3-Embedding-4B for embedding representation. As shown in Figure~\ref{fig:density}, the generated \rectag{<rec>}...\rectag{</rec>} fields align markedly closer with the golden skill descriptions in the latent representation space, validating the advantage of the generative skill retrieval mechanism.

\textbf{The effect of warm start}.
In \modelname{}, we introduce an instruction tuning stage as a warm start prior to GRPO. To analyze the efficacy of this design in Table~\ref{tab:ablation}, Figure~\ref{fig:rl_nowarm} compares the GRPO training reward curves with and without the warm-start phase. As illustrated, the warm-started model initiates training with a significantly higher reward compared to the cold-start version and ultimately converges to a higher performance. Intuitively, the warm-start phase can teach the model to generate well-structured, high-quality \rectag{<rec>} and \anstag{<answer>} fields. This prior knowledge facilitates successful trajectory rollouts during GRPO, thereby providing more informative training gradients to drive policy optimization.

\subsection{Robustness Analysis}

\begin{table}[t]
    \centering
    \caption{Performance on OOD datasets}
    \vspace{-0.2cm}
    \renewcommand{\arraystretch}{1.05}
    \resizebox{0.4\textwidth}{!}{
    \begin{tabular}{c|cc}
        \toprule
        \textbf{Method}    & \textbf{MBPP+} & \textbf{LiveCodeBench} \\ \midrule
        Base (Qwen3-4B)  & 55.82     & 17.96   \\
        \modelname{}   & 62.17       & 19.16     \\ \bottomrule
    \end{tabular}}
    \label{tab:OOD}
    \vspace{-0.1cm}
\end{table}

\textbf{Robustness to datasets}. To evaluate the robustness of \modelname{} on out-of-distribution datasets, we further conduct experiments on MBPP+~\cite{mbpp+} and LiveCodeBench~\cite{livecodebench}. Specifically, we construct a new retrieval library using skills collected from SkillRet~\cite{skillret}, which comprises approximately 11,085 skills centered on coding and software engineering. In particular, these skills are entirely unseen during our training phase. Since these skills' domain-specific coding knowledge is highly aligned with the target tasks, they can offer valuable guidance for problem-solving in both benchmarks. As presented in Table~\ref{tab:OOD}, \modelname{} consistently provides notable performance gains for the base model across these OOD benchmarks, demonstrating its robust generalization capabilities in both generative skill retrieval and skill utilization under unseen contexts.

\begin{figure}[t]
  \centering
  \begin{subfigure}[b]{0.48\columnwidth}
    \includegraphics[width=\textwidth]{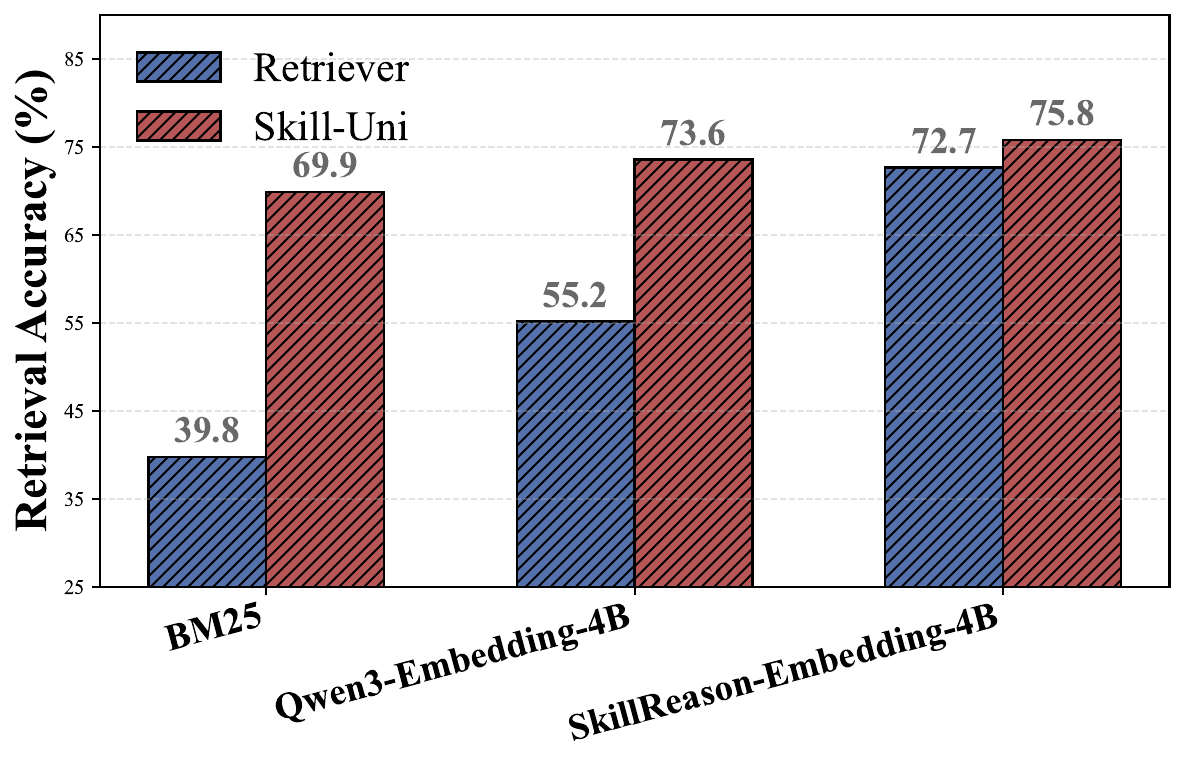}
    \caption{The retrieval accuracy of \modelname{} with different retrieval engines}
    \label{fig:retrieval_acc}
  \end{subfigure}
  \hfill
  \begin{subfigure}[b]{0.48\columnwidth}
    \includegraphics[width=\textwidth]{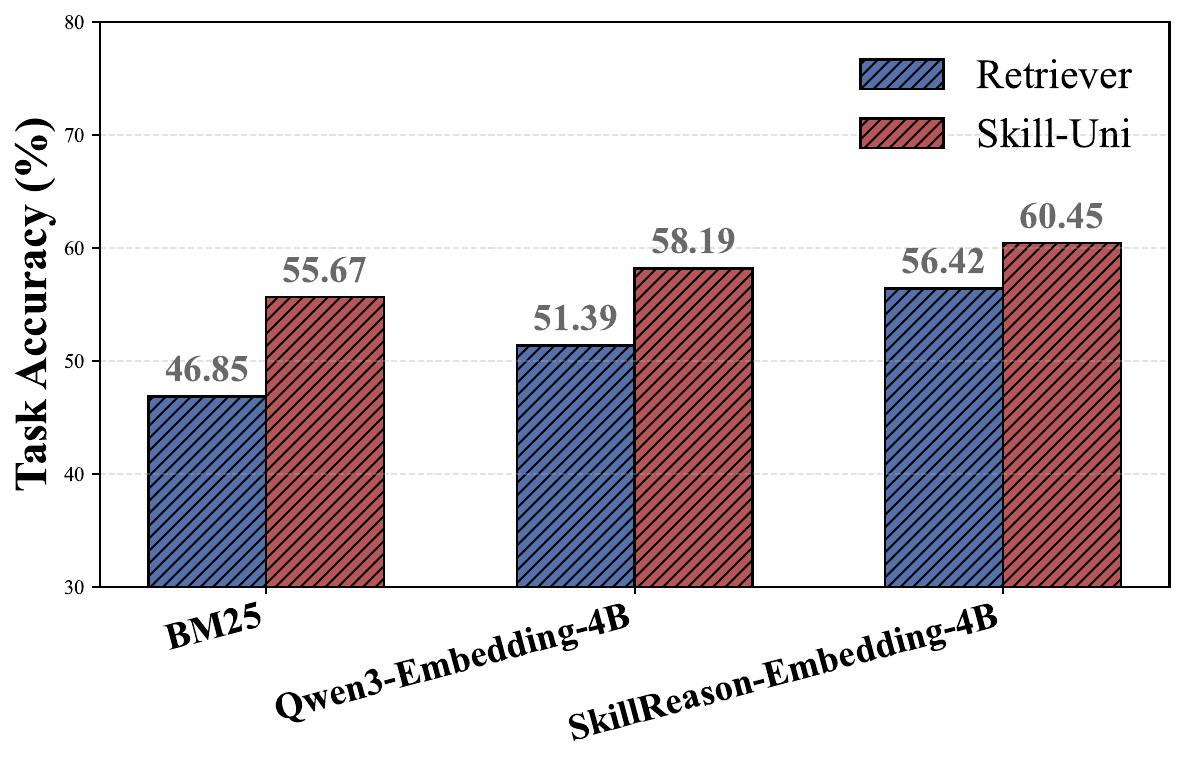}
    \caption{The downstream task accuracy of \modelname{} with different retrieval engines}
    \label{fig:task_acc}
  \end{subfigure}
  \caption{The performance analysis of \modelname{} applied with different retrieval methods. The base model is Qwen3-4B.}
  \vspace{-0.2cm}
  \label{fig:retriever}
\end{figure}

\textbf{Robustness to Retrieval Engines}. A key advantage of \modelname{} is that its generative skill retrieval mechanism is agnostic to the underlying retrieval engine. To evaluate this property, we pair \modelname{} with several retrieval methods, including BM25, Qwen3-Embedding-4B, and SkillReason-Embedding-4B. As shown in Figure~\ref{fig:retrieval_acc} and Figure~\ref{fig:task_acc}, which illustrate the results on retrieval accuracy and downstream task accuracy, respectively, three observations emerge: 1) Skill-specific retrievers, such as SkillReason-Embedding-4B, generally provide stronger retrieval performance than more generic alternatives. 2) Across different retrieval backends, \modelname{} consistently improves skill retrieval performance, highlighting the robustness and compatibility of our proposed generative skill retrieval formulation. 3) These retrieval gains, together with improved skill utilization, further translate into consistent improvements in downstream task accuracy across all three retrivers.

\subsection{Case Study}

To better understand the effectiveness of \modelname{}, we provide two representative cases in this part, in which \modelname{} derives correct solutions due to its remarkable capability in both skill retrieval and utilization. Qwen3-4B is set as the base model here.

\begin{figure}[t]
  \centering
  \includegraphics[width=0.92\columnwidth]{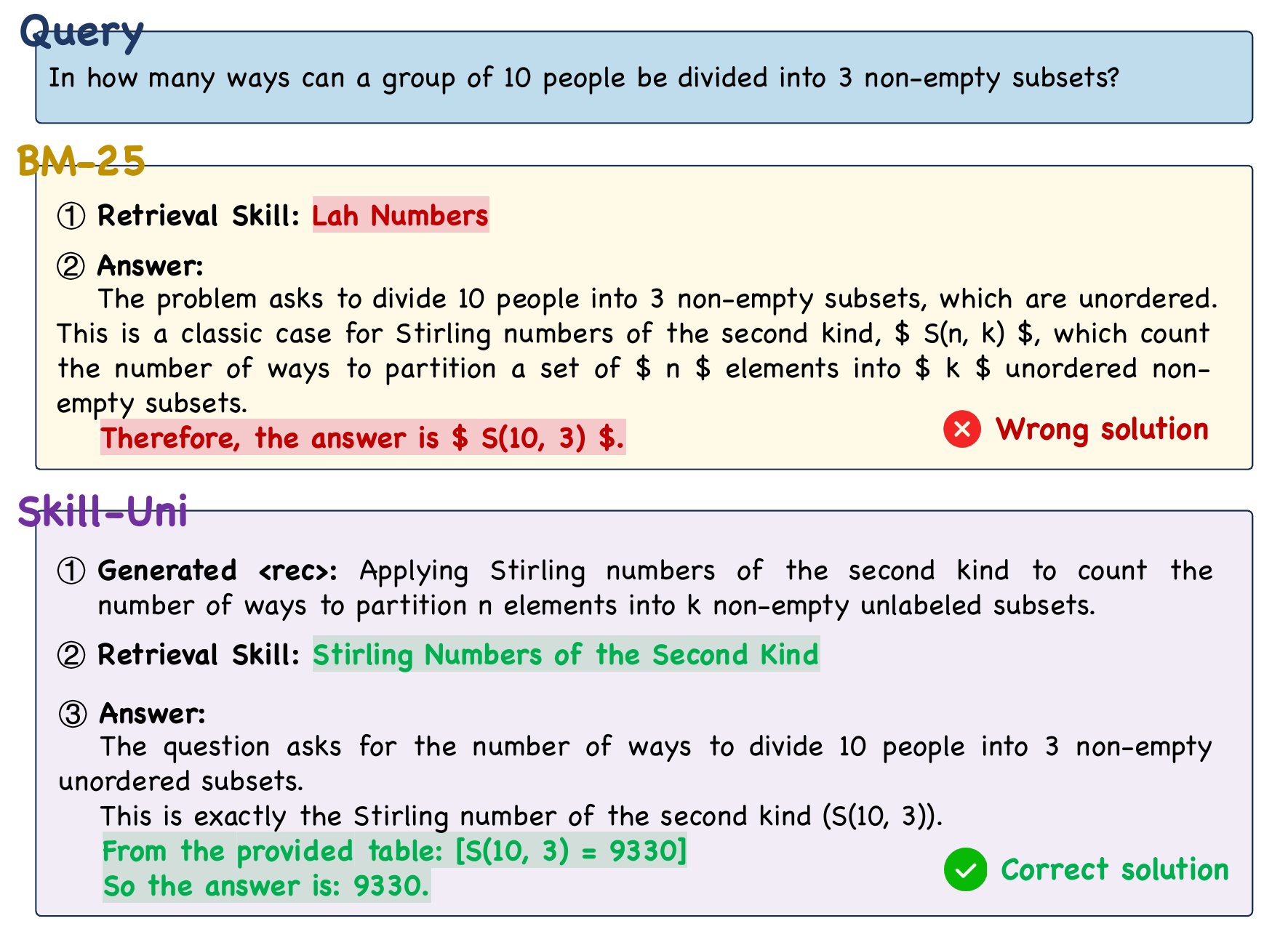}
  \caption{A case in TheoremQA where \modelname{} retrieves the correct skill while BM25 fails.}
  \vspace{-0.1cm}
  \label{fig:case1}
\end{figure}

\textbf{Case on skill retrieval}. As shown in Figure~\ref{fig:case1}, this case demonstrates the decisive role of retrieval precision in enabling successful downstream reasoning. In this task, \modelname{} employs its generative retrieval mechanism to proactively produce a clear desired skill description, explicitly referencing \textit{Stirling numbers of the second kind} for partitioning unlabeled subsets. This semantic representation aligns closely with the ground-truth skill, allowing the model to retrieve a skill document containing explicit precomputed value tables ($S(10, 3) = 9330$) and directly extract the correct numerical answer. Conversely, BM25 relies solely on surface-level term matching from the original prompt. Due to lexical overlap regarding terms like "partition" and "non-empty subsets", BM25 incorrectly retrieves the \textit{Lah Numbers} skill —an algorithmically irrelevant document addressing ordered list partitions. Lacking the necessary Stirling lookup table, the model can only produce a symbolic expression $S(10, 3)$ rather than a concrete numerical result.

\begin{figure}[t]
  \centering
  \includegraphics[width=0.92\columnwidth]{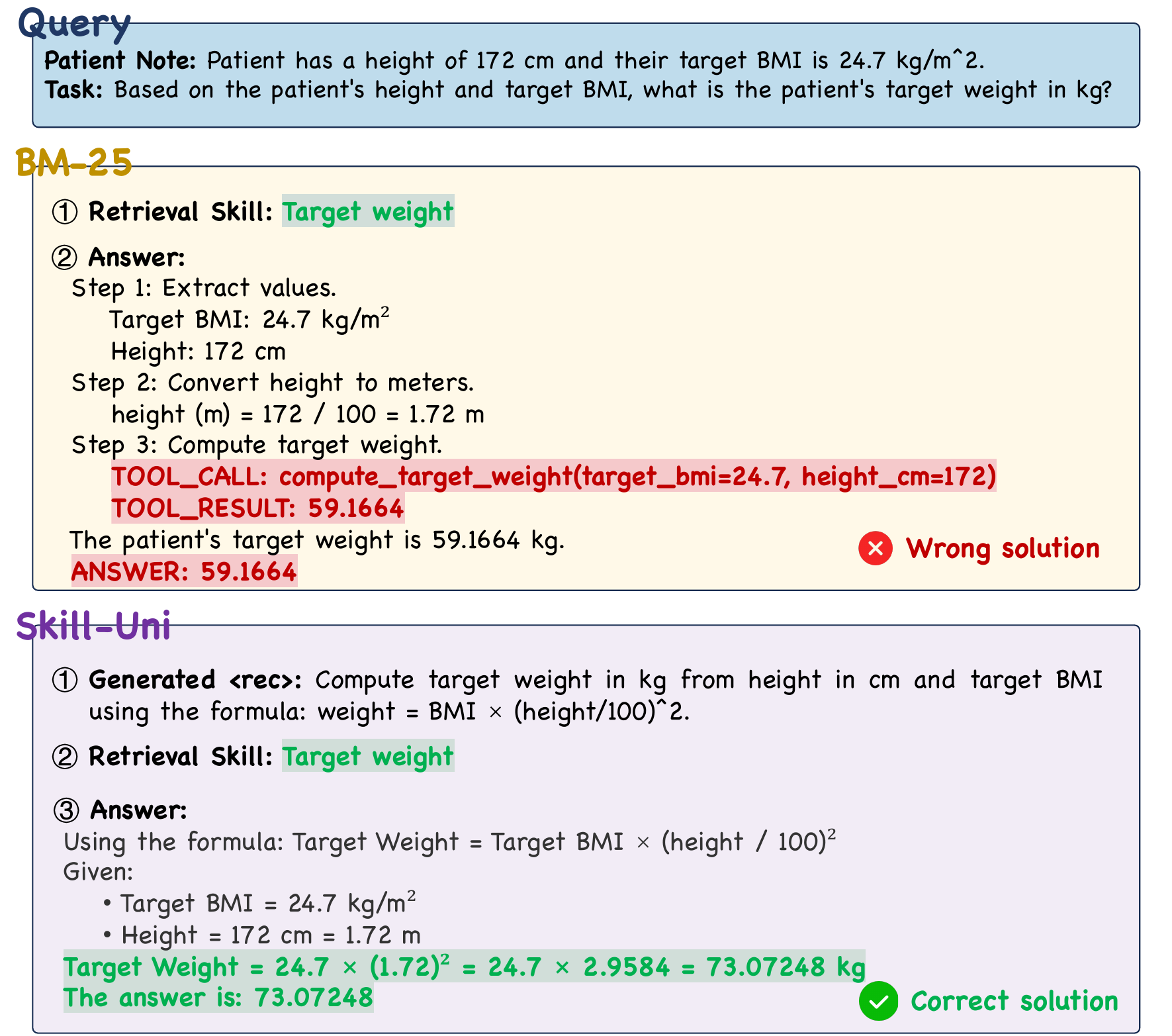}
  \caption{A case in MedCalc-Bench where both BM25 and \modelname{} retrieve the golden skills, while only \modelname{} derives the correct solution.}
  \label{fig:case2}
  \vspace{-0.1cm}
\end{figure}

\textbf{Case on skill utilization}. As shown in Figure~\ref{fig:case2}, in this case, both \modelname{} and BM25 successfully retrieve the identical ground-truth skill \textit{Target weight} for subsequent question solution. However, their downstream utilization behaviors diverge markedly. Specifically, \modelname{} demonstrates genuine functional execution by faithfully parsing the target formula ($\text{Weight} = \text{BMI} \times \text{height}^2$) from the skill documentation and performing the arithmetic computation to yield the exact answer ($73.07\text{ kg}$). In contrast, BM25 succumbs to superficial pattern imitation: rather than executing the underlying mathematical logic, it only formally copies the \texttt{TOOL\_CALL} and \texttt{TOOL\_RESULT} template provided in the skill’s example and hallucinates an incorrect output ($59.1664$). This contrast verifies \modelname{}'s superior capability in skill comprehension and utilization.

%% file: conclusion.tex
\section{Conclusion}

In this paper, we propose \modelname{}, a unified reinforcement learning framework that jointly optimizes skill retrieval and utilization for skill-augmented generation. By adopting a generative skill retrieval paradigm, \modelname{} unifies these two processes within a single model, overcoming the limitations of conventional approaches that optimize them separately. To effectively train this framework, we introduce a two-stage training strategy combining warm-up instruction tuning and GRPO, enabling the model to improve both skill retrieval and downstream task-solving capabilities. Extensive experiments on five skill-related benchmarks show that \modelname{} achieves state-of-the-art performance, with the highest skill retrieval accuracy and strong compatibility with different retrieval engines. In-depth model analysis also highlights the importance of the warm-up stage and generative skill retrieval.